\documentclass[11pt]{article}

\usepackage[preprint]{acl}

\usepackage{times}
\usepackage{latexsym}

\usepackage[T1]{fontenc}

\usepackage[utf8]{inputenc}

\usepackage{microtype}

\usepackage{inconsolata}

\usepackage{graphicx}

\usepackage{multirow}
\usepackage{booktabs}
\usepackage{amssymb}
\usepackage{subfigure}
\usepackage{amsmath}
\usepackage{algorithm}
\usepackage{algorithmic}
\usepackage{subcaption}

\usepackage{color}

\title{DOne: Decoupling Structure and Rendering for High-Fidelity Design-to-Code Generation}

\author{
Xinhao Huang$^{1}$,
Jinke Yu$^{2}$,
Wenhao Xu$^{2}$,
Zeyi Wen$^{1,3}$\textsuperscript{*},
Ying Zhou$^{4}$,
Junzhuo Liu$^{5}$,
Junhao Ji$^{2}$,
Zulong Chen$^{2}$\thanks{~~Corresponding Author} \\
\textsuperscript{\rm1}HKUST (Guangzhou), Guangzhou, China \quad
\textsuperscript{\rm2}Alibaba Group, Hangzhou, China~~~ \\
\textsuperscript{\rm3}HKUST, Hong Kong, China~~~ \quad
\textsuperscript{\rm4}University of Electronic Science, Chengdu, China~~~ \\
\textsuperscript{\rm5}Zhejiang Lab, Hangzhou, China~~~
\\
\texttt{chenzulong198867@gmail.com, wenzeyi@ust.hk}
}

\begin{document}
\maketitle
\begin{abstract}
While Vision Language Models (VLMs) have shown promise in Design-to-Code generation, they suffer from a ``holistic bottleneck''—failing to reconcile high-level structural hierarchy with fine-grained visual details, often resulting in layout distortions or generic placeholders. To bridge this gap, we propose DOne, an end-to-end framework that decouples structure understanding from element rendering. DOne introduces (1) a learned layout segmentation module to decompose complex designs, avoiding the limitations of heuristic cropping; (2) a specialized hybrid element retriever to handle the extreme aspect ratios and densities of UI components; and (3) a schema-guided generation paradigm that bridges layout and code. To rigorously assess performance, we introduce HiFi2Code, a benchmark featuring significantly higher layout complexity than existing datasets. Extensive evaluations on the HiFi2Code demonstrate that DOne outperforms exiting methods in both high-level visual similarity (e.g., over 10\% in GPT Score) and fine-grained element alignment. Human evaluations confirm a $3\times$ productivity gain with higher visual fidelity.
\end{abstract}

\section{Introduction}

The transformation of visual design drafts into executable code—known as Design2Code—is a pivotal yet labor-intensive process in web development. By automating the conversion of visual prototypes into HTML and CSS, this technology reduces technical barriers for non-experts \cite{REMAUI} and allows developers to focus on aesthetic innovation rather than repetitive implementation \cite{moran2018machine_49, chen2018ui_13, moran2018automated_50, wu2021screen_68, DCGen, Design2Code, Web2Code}. 

While early approaches utilized heuristics or CNN/LSTM-based architectures \cite{chen2020object, chen2022towards, abdelhamid2020deep_5, pix2code, acsirouglu2019automatic, chen2022code, xu2021image2emmet}, they were often limited to simplistic UIs or specific mobile domains, suffering from representational bias in synthetic training data. Recently, Vision Language Models (VLMs) have marked a paradigm shift \cite{gpt4v, hou2024large_36, dong2024self_25}. However, despite these advancements, applying VLMs directly to Design2Code often yields suboptimal outcomes \cite{pix2code, Sketch2code}.

\begin{figure}[t]
    \centering

    \subfigure[Original reference.] { 
        \centering
        \includegraphics[height=0.11\textheight]{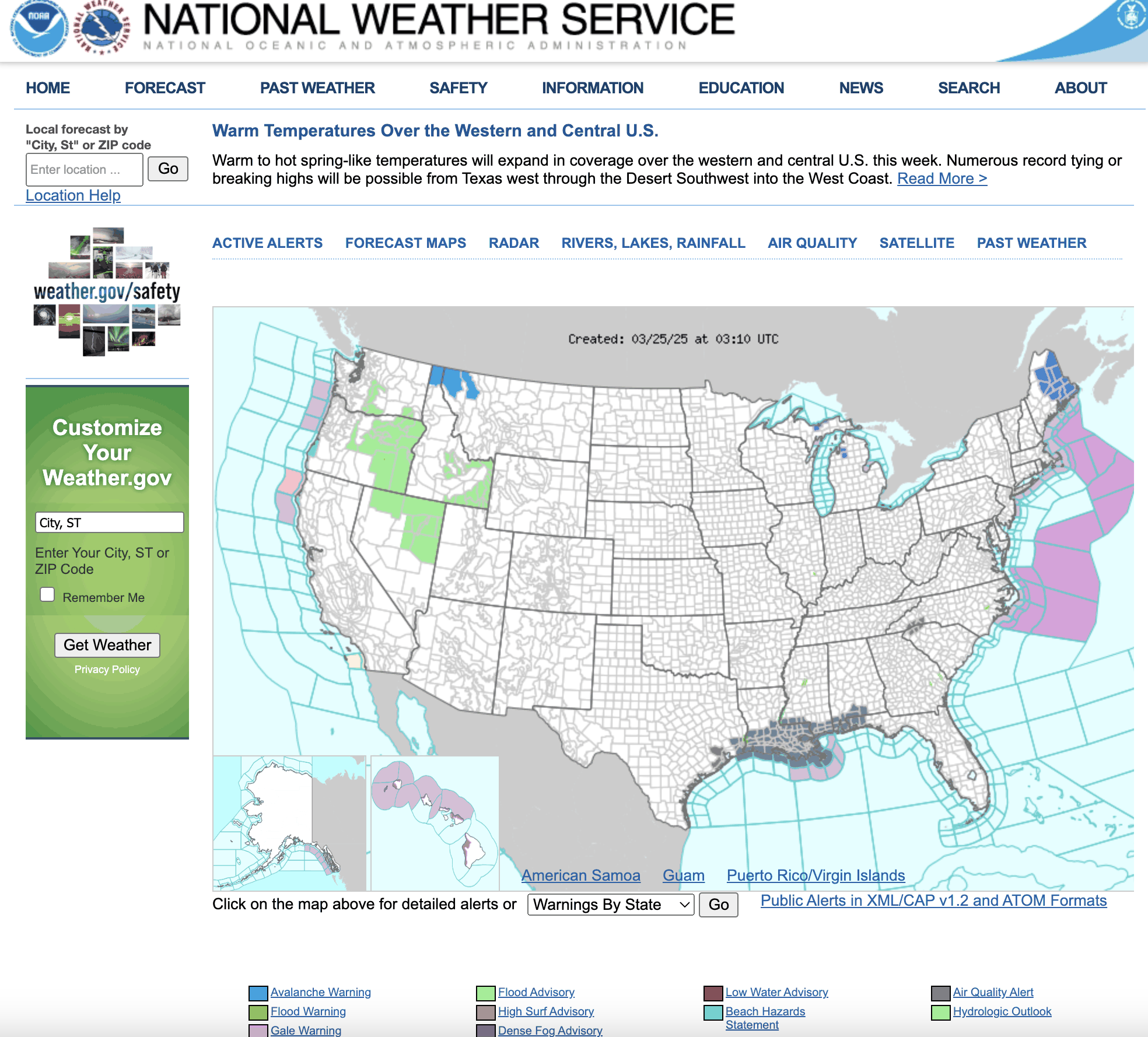}
        \label{fig:comparison_1}
    }
    \hfill
    \subfigure[Output by baseline.]{
        \centering
        \includegraphics[height=0.11\textheight]{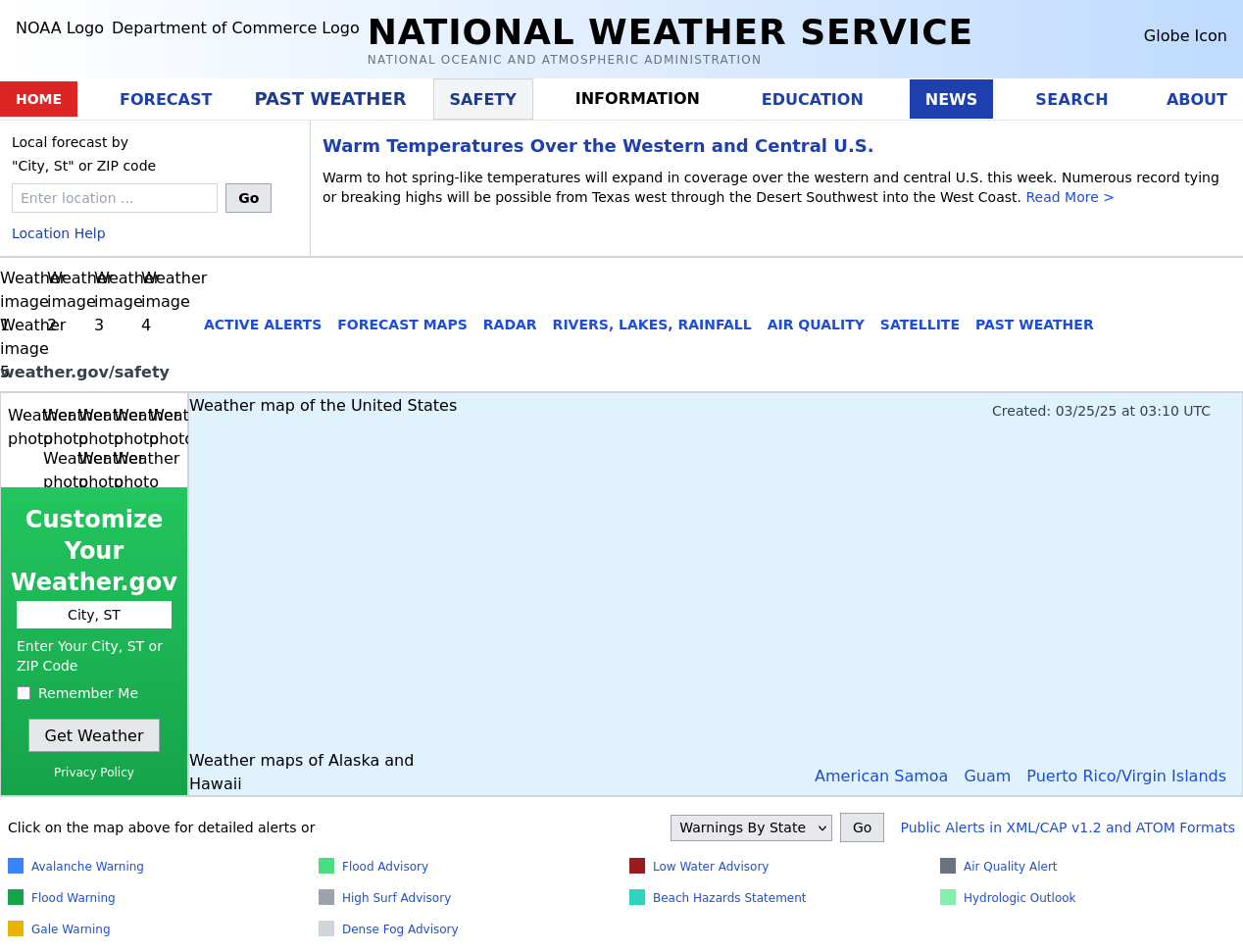}
        \label{fig:comparison_2}
    }
    
    \subfigure[Erroneous layout parsing.] {
        \centering
        \includegraphics[width=0.46\linewidth]{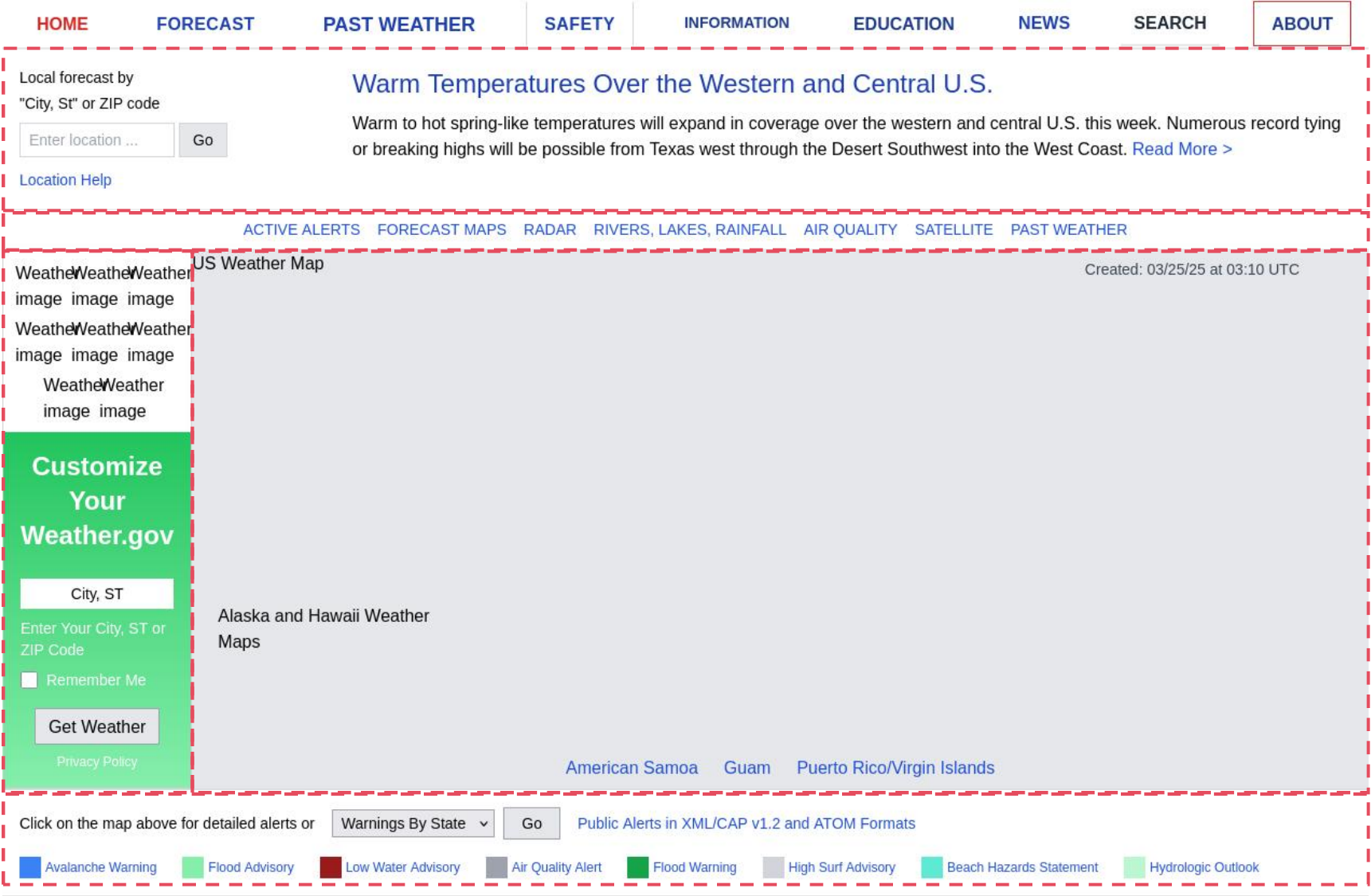}
        \label{fig:error_1}
    }
    \hfill
    \subfigure[Loss of visual element.] {
        \centering
        \includegraphics[width=0.46\linewidth]{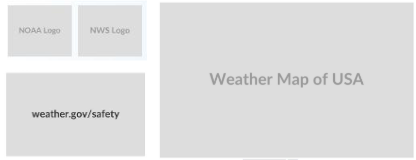}
        \label{fig:error_2}
    }
    \caption{Reconstruction comparison (a-b) and illustrative examples of baseline failures (c-d).}
    \label{fig:comparison}
\end{figure}

Our preliminary investigation reveals that current state-of-the-art methods continue to struggle with balancing high-level structure recovery and fine-grained visual reconstruction. As illustrated in Figure~\ref{fig:comparison_1} and~\ref{fig:comparison_2}, when processing complex real-world designs, baselines such as S2C, CoT2C, and even the recent divide-and-conquer approach DCGen \cite{DCGen} frequently exhibit layout distortions and omit critical elements. Specifically, we identify three interrelated failure modes that compromise fidelity:
(1) Erroneous Hierarchy Parsing \cite{gui2024vision2ui, wan2024mrweb}: Models often misinterpret the semantic relationships among sections (e.g., confusing sidebars with content blocks), resulting in flawed document structures (see Figure~\ref{fig:error_1}). 
(2) Loss of Fine-Grained Visual Elements \cite{Design2Code}: Discrete assets like icons and logos are often replaced by generic placeholders or ignored entirely, diminishing visual authenticity (see Figure~\ref{fig:error_2}).
(3) Holistic Processing Overload: Processing a complex page as a monolithic image exceeds the capacity of current VLMs, leading to a trade-off where structural integrity is sacrificed for local details.

To address these limitations, we propose DOne, a novel end-to-end framework that decouples structural understanding from element rendering. Unlike methods that rely on heuristic cropping or direct generation, DOne adopts a hierarchical strategy. We first construct WebSeg, a large-scale dataset with detailed layout annotations, to train a specialized learned layout segmentation model. This decomposes the design into semantically meaningful blocks, breaking the ``holistic bottleneck.'' Concurrently, a visual element retriever identifies and preserves fine-grained assets (e.g., icons, buttons). Finally, DOne utilizes a schema-guided generation process: it infers a hierarchical JSON blueprint from the segmented layout and detected elements, which then guides the VLM to synthesize high-fidelity, element-aware UI code.

To support rigorous evaluation, we introduce HiFi2Code, a benchmark dataset featuring significantly greater layout complexity and element density than existing resources like Design2Code \cite{Design2Code} and Web2Code \cite{Web2Code}. Extensive evaluations on HiFi2Code demonstrate that DOne raises the CLIP Score to 0.7435 and outperforms state-of-the-art baselines by over 10\% in GPT Score. Human evaluations further confirm that DOne achieves a threefold improvement in productivity while generating code that is visually faithful to the original design.

In summary, the key contributions of this study are:
\begin{itemize}
    \item We propose DOne, a hierarchical framework unifying learned layout segmentation, fine-grained element retrieval, and schema-guided generation.
    \item We release WebSeg (for training) and HiFi2Code (for evaluation) to address data scarcity in complex web UI tasks.
    \item We demonstrate through extensive experiments that DOne significantly outperforms existing methods in both visual fidelity and developer productivity.
\end{itemize}

\section{Related Work}
\subsection{Traditional Vision-Based Approaches}
Early Design2Code approaches primarily relied on computer vision techniques, particularly optical character recognition, to identify UI elements, and employed heuristic rules to generate UI code~\cite{luo2021survey, archana2024deep, REMAUI}. Subsequent CNN-based methods demonstrated significant advancements~\cite{pix2code, acsirouglu2019automatic, chen2022code, xu2021image2emmet}. Notably, pix2code~\cite{pix2code} integrated CNNs with RNNs to generate domain-specific language code from GUI screenshots. Another method leverages CNN to understand UI elements and their spatial layout within images, and further extracts visual features from UI images, thereby transforming a UI design mockup into a GUI skeleton \cite{chen2018ui_13}. Andre et al. \cite{CizottoSMC23} investigated the application of weakly-supervised semantic segmentation to validate the effectiveness of deep neural networks in the segmentation and classification of web elements. Similarly, image2emmet~\cite{xu2021image2emmet} directly transformed images into HTML code. However, these methods remain limited to simplistic UI designs, struggling with complex layouts and context-aware code generation, rendering them inadequate for developing real-world websites.

\subsection{VLM-Driven Methods}
The emergence of VLMs has revolutionized context-aware code generation \cite{gui2024vision2ui, Web2Code, DCGen, zhou2024bridging, wan2024mrweb, hou2024large_36, dong2024self_25} and fine-grained visual understanding \cite{gpt4v, YuSRZZMLLWX24_73, NijkampPHTWZSX23_53, ArakelyanDMR23_9, MastropaoloPB22_46}. Prior works have optimized performance through prompt engineering \cite{Design2Code}, parsing strategies \cite{lee2023pix2struct}, and interactive modeling \cite{xiao2024interaction2code_69, zhou2024bridging}. While structural methods leverage layout trees \cite{UICopilot, layoutcoder, screencoder} or heuristic cropping \cite{DCGen}, they typically depend on DOM parsing or heuristics. However, heuristic segmentation lacks semantic understanding, often leading to misaligned layouts. In contrast, DOne employs a learned layout segmentation model and a dedicated element retriever to decouple structural generation from fine-grained rendering, ensuring high fidelity in complex scenarios.

\subsection{Benchmark Datasets for Design2Code}
Existing datasets range from synthetic collections of simple HTML elements \cite{soselia2023learning, laurenccon2024unlocking, pix2code} to instruction-tuning corpora like Web2Code \cite{Web2Code}. While Design2Code \cite{Design2Code} introduced real-world webpages, these datasets often lack the high element density and structural nesting typical of modern interfaces. This limitation hinders the evaluation of production-ready code generation, motivating the visually rich HiFi2Code benchmark.

\begin{figure*}[t]
    \centering
    \includegraphics[width=0.99\textwidth]{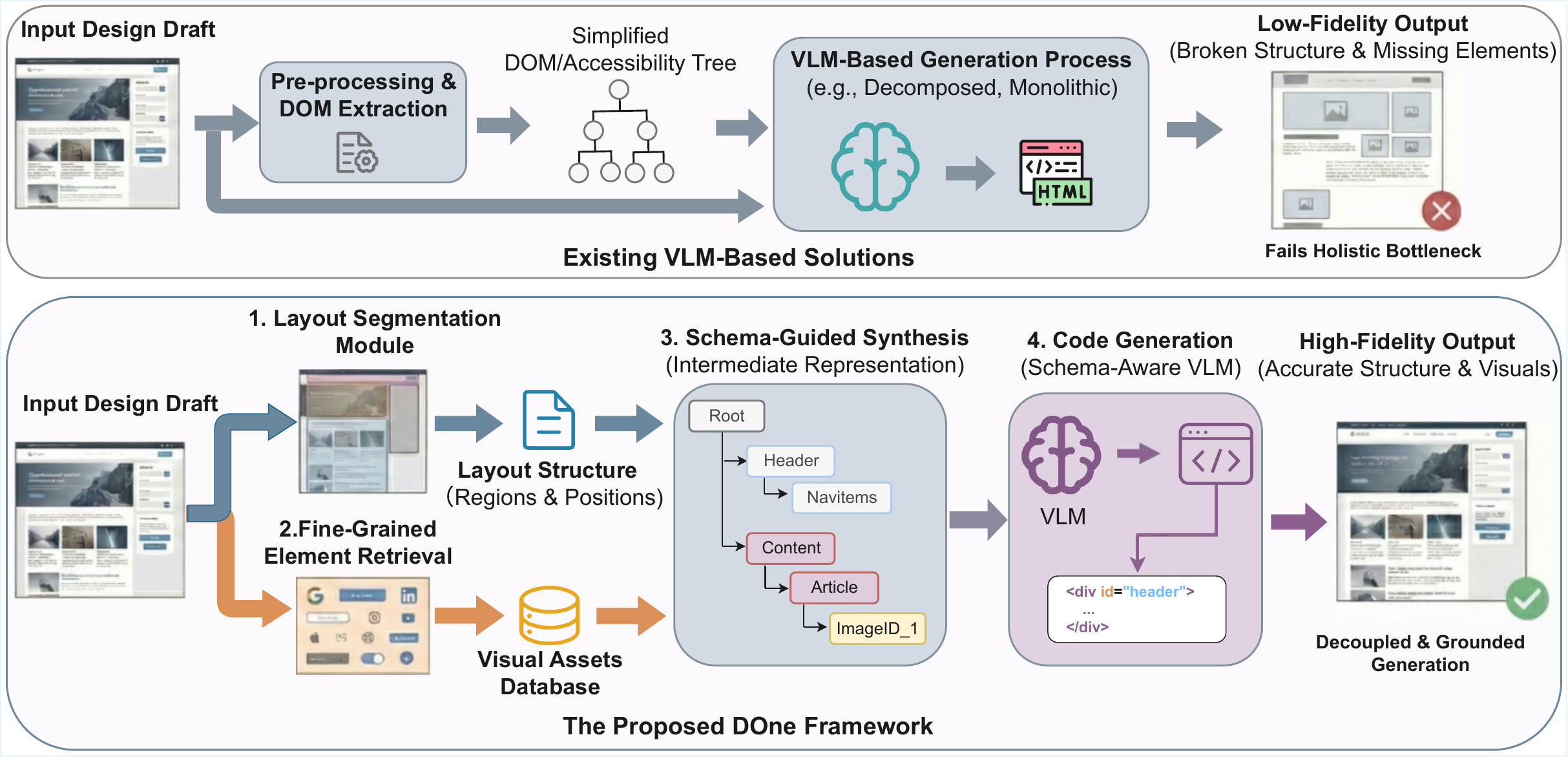}
    \caption{The DOne Framework. DOne employs layout segmentation and visual element detection, followed by schema-guided code generation to ensure structural and visual fidelity.}
    \label{fig:pipeline}
\end{figure*}

\section{Dataset Construction}
\label{sec:datasets}

To address the scarcity of high-fidelity training and evaluation data, we introduce two datasets: WebSeg for training layout segmentation and HiFi2Code for end-to-end evaluation.

\begin{table}[b]
    \small
    \centering
    \begin{tabular}{lccc}
        \toprule
        Metric & Min & Max & Avg \\
        \midrule
        Boxes/page & 2 & 7 & 3.53 \\
        Elements/page & 1 & 64 & 4.66 \\
        Elements/box & 0 & 16 & 1.11 \\
        Tokens/box & 0 & 1518 & 69.97 \\
        Box-to-page Ratio & 0\% & 99\% & 25\% \\
      \bottomrule
    \end{tabular}
    \caption{Statistics of WebSeg.}
    \label{tab:layout}
\end{table}

\begin{table}[b]
    \centering
    \resizebox{\columnwidth}{!}{
        \begin{tabular}{lccc}
            \toprule
            Metric & Design2Code & Web2Code & \textbf{HiFi2Code} \\
            \midrule
            Avg. boxes/page & 3.88 & 2.84 & \textbf{5.32} \\
            Avg. elements/page & 2.59 & 1.43 & \textbf{14.39} \\
            Avg. elements/box & 0.53 & 0.51 & \textbf{2.80} \\
            Element coverage & 81.4\% & 66.9\% & \textbf{100\%} \\
            Large elements & 20.7\% & 17.7\% & \textbf{44.2\%} \\
            \bottomrule
        \end{tabular}
    }
    \caption{Comparison of evaluation datasets.}
    \label{tab:eval}
\end{table}

\subsection{WebSeg} 
We curate 30,000 design drafts from open-source repositories \cite{kaggle}. Expert UI/UX designers annotated these images with semantic regions (e.g., headers, navbars) under strict guidelines: (1) maximal coverage, (2) bounding boxes $\leq$ 300 tokens, and (3) non-nested structures to ensure clarity.

As summarized in Table~\ref{tab:layout}, the dataset exhibits considerable diversity and complexity, capturing fine-grained layout structures. On average, each page contains 3.53 boxes (max: 7), and 4.66 visual elements (range: 1–64), indicating substantial visual richness. Each box contains up to 16 visual elements (avg. 1.11) and 1518 tokens (avg. 69.97), reflecting varied content density. The box-to-page area ratio ranges from 0 to 99.82\% (avg. 24.98\%), further underscoring layout diversity. 

\subsection{HiFi2Code} 
Existing benchmarks often emphasize text over visual richness \cite{Design2Code, Web2Code}. To better reflect practical high-fidelity scenarios, we construct HiFi2Code, a dataset comprising 200 modern webpages \cite{hao123, kazada}. The sample size aligns with the standard evaluation subsets used in recent studies, yet our dataset is explicitly curated for enhanced visual and structural diversity.

As detailed in Table~\ref{tab:eval}, HiFi2Code exhibits superior structural and visual complexity. It features a deeper hierarchy with 5.32 sub-layouts (vs. 3.88 in Design2Code) and 14.39 visual elements per page (vs. 2.59). Moreover, with 2.80 elements per box and 100\% coverage, HiFi2Code presents a tighter visual-layout coupling and a higher prevalence of large-scale assets (44.2\%), establishing a rigorous benchmark for high-fidelity reconstruction.

\section{Methodology}

The DOne framework (Figure~\ref{fig:pipeline}) adopts a hierarchical strategy consisting of three stages: (1) Layout Segmentation to decompose the page; (2) Visual Element Retrieval to capture fine-grained assets; and (3) Schema-Guided Generation to synthesize the final code. 

\begin{figure}[t]
    \centering
    \includegraphics[width=0.9\linewidth]{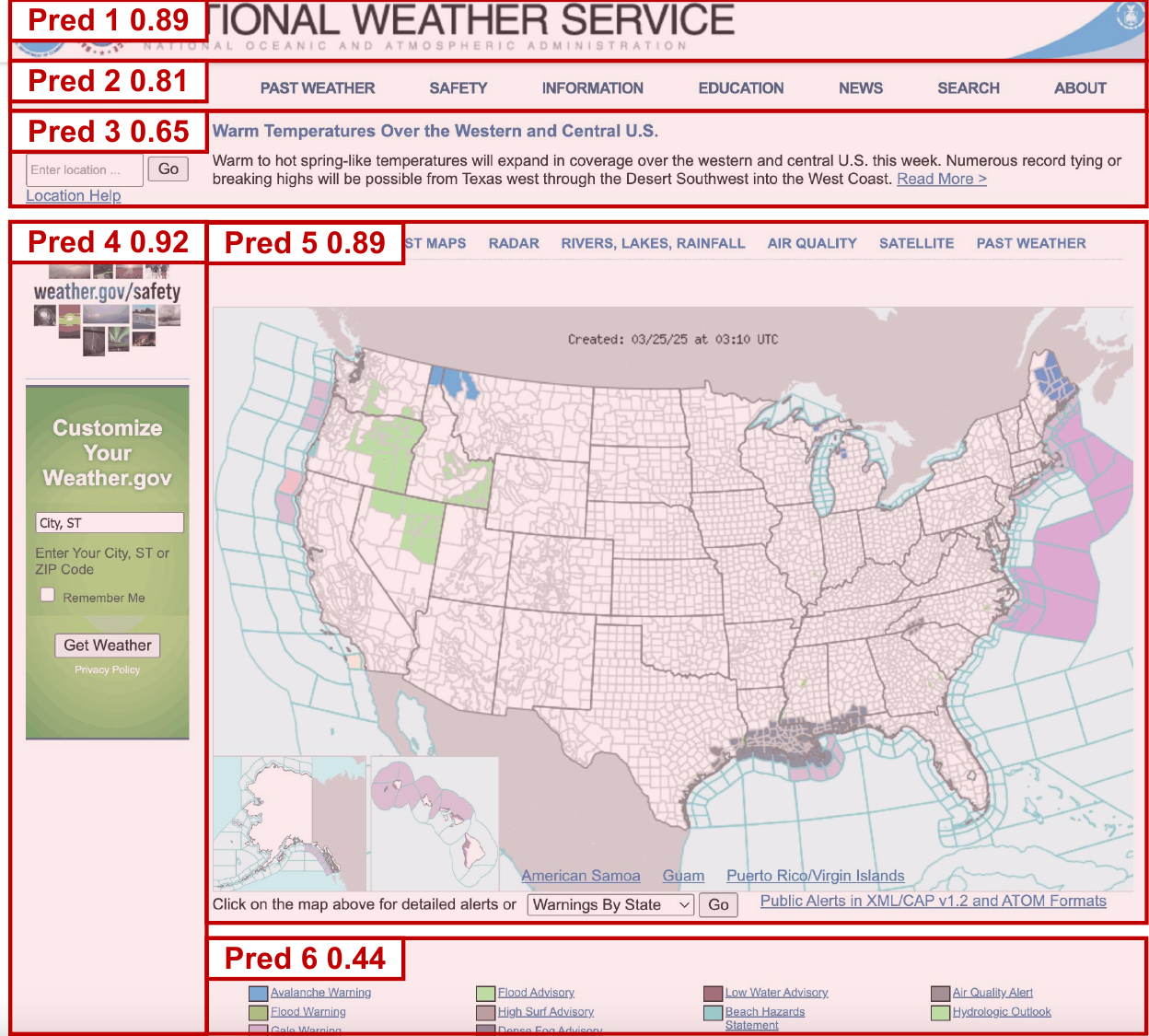}
    \caption{Example of layout segmentation.} 
    \label{fig:segment_example}
\end{figure}

\paragraph{Concept of Design2Code} The task of Design2Code aims to automatically translate visual reference images (e.g., screenshots or mockups) into executable frontend code, typically HTML and CSS. While Vision Language Models (VLMs) have demonstrated potential in visual understanding, they often struggle to generate code that faithfully reflects intricate designs due to high visual complexity and a scarcity of fine-grained annotated data. To address these challenges, we propose an end-to-end framework designed to decouple structural understanding from element rendering, ensuring high-fidelity reconstruction.

\subsection{Webpage Layout Segmentation}
\label{sec:layout_section}

To bridge the gap between holistic visual understanding and fine-grained code generation, we first decompose the full-page screenshot into semantically meaningful sub-regions (e.g., headers, sidebars). This mimics the macroscopic planning phase of human designers, ensuring structural coherence.

\paragraph{Segmentation Model.}
We employ RT-DETR (Real-Time DEtection TRansformer) \cite{rtdetr}, an end-to-end transformer-based detector chosen for its ability to predict bounding boxes. Formally, given an input image $I$, the model predicts a set of bounding boxes $B$. 
The detection probability is formulated as:
\begin{equation}
\resizebox{1.0\columnwidth}{!}{$
P_{RT\text{-}DETR}(B \mid I) = \prod_{i=1}^{N} \sigma(f_{cls}(d_i)) \cdot \mathcal{B}(b_i; f_{reg}(d_i)),
$}
\end{equation}
where $d_i$ represents the object query feature refined by the encoder-decoder architecture, and $\mathcal{B}$ denotes the regressed box coordinates.

We train the RT-DETR on our annotated dataset for 15 epochs using SGD (lr=0.01, momentum=0.937) with input images resized to $720 \times 720$ pixels. Experimental results show that the model achieves an mAP50 of 0.820. This high metric indicates that DOne effectively identifies and localizes over 80\% of the significant layout structures with high precision, ensuring that the subsequent code generation is grounded in a complete and accurate structural framework.

\paragraph{Region Box Optimization.} 

Raw model predictions occasionally yield slightly overlapping regions due to ambiguous visual boundaries. To ensure structural coherence, we implement a post-processing algorithm. As outlined in the Algorithm \ref{alg:box_optimization}: first, all detected boxes are sorted in ascending order by area; then, each box is iteratively compared with currently retained boxes by calculating the Intersection over Union (IoU). If the overlap exceeds a predefined threshold, the confidence scores are further compared—only boxes with significantly higher confidence (e.g., exceeding by a factor of 1.2) are retained, otherwise redundant boxes are discarded. Empirical observations indicate that the method is robust to minor variations in the IoU threshold ($[1.1, 1.3]$), with $1.2$ yielding optimal separation for dense layouts. As shown in Figure \ref{fig:segment_example}, this results in a clean, non-overlapping segmentation of complex pages.

\begin{algorithm}[t]
\caption{Layout Region Box Optimization}
\label{alg:box_optimization}
\small
\begin{algorithmic}[1]
    \STATE Sort detected boxes by region area (ascending)
    \STATE Initialize optimized\_box $\leftarrow$ []
    \FOR{each box in sorted\_boxes}
        \STATE should\_keep $\leftarrow$ True
        \FOR{each kept\_box in optimized\_boxes}
            \STATE iou = CalculateIoU(box, kept\_box)
            \IF{iou $>$ threshold}
                \IF{box.score $>$ kept\_box.score $\times$ 1.2}
                    \STATE Remove kept\_box 
                \ELSE
                    \STATE should\_keep $\leftarrow$ False
                    \STATE break
                \ENDIF
            \ENDIF
        \ENDFOR
        \IF{should\_keep}
            \STATE Add box to optimized\_boxes
        \ENDIF
    \ENDFOR
\end{algorithmic}
\end{algorithm}

\subsection{Fine-Grained Visual Element Detection}
\label{sect:element_section}

Unlike general object detection, identifying Web UI elements presents unique challenges: (1) Extreme Aspect Ratios, where elements like navigation bars exhibit ratios (e.g., $1:20$) rarely seen in natural images; (2) High Density, particularly for small icons in footers or grids; and (3) Semantic Gap, where functional widgets (e.g., `Input Field') differ fundamentally from generic objects.

To address these issues, we propose a hybrid detection strategy that synergizes  DETR \cite{detr} and YOLOv10 \cite{yolov10}.  We exploit DETR's attention mechanism to resolve layout-defining structures through global context, while deploying YOLOv10 to handle high-density, small elements via efficient anchor-based detection.

\paragraph{Hybrid Formulation.}
Given input image $I$, the detection probabilities for DETR and YOLO are formulated respectively as:
\begin{equation}
\resizebox{1.0\columnwidth}{!}{$
P_{DETR} = \prod \sigma(f_{cls}(h_i)) \cdot \mathcal{N}(b_i; f_{reg}(h_i), \Sigma)
$}
\end{equation}
\begin{equation}
\resizebox{1.0\columnwidth}{!}{$
P_{YOLO} = \prod P(obj) \cdot P(cls \mid obj) \cdot P(bbox \mid obj)
$}
\end{equation}

We define a binary class selection matrix $C \in \{0, 1\}^{K}$, where $K$ is the number of element categories. This matrix is manually constructed based on the statistical properties of UI elements: $C_k = 1$ (assigning to DETR) for large, context-dependent classes (e.g., Navbar, Sidebar, Hero Image), and $C_k = 0$ (assigning to YOLO) for small, dense elements (e.g., Icons, Buttons, Text Labels). 
Given input image $I$, the final detection set $\hat{B}$ is obtained by selecting predictions based on the category of the detected object:
\begin{equation}
\resizebox{1.0\columnwidth}{!}{$
\hat{B} = \{ b \mid (b \in P_{DETR} \land C_{cls(b)}=1) \lor (b \in P_{YOLO} \land C_{cls(b)}=0) \}
$}
\end{equation}

\begin{figure}[t]
    \centering
    \includegraphics[width=0.9\linewidth]{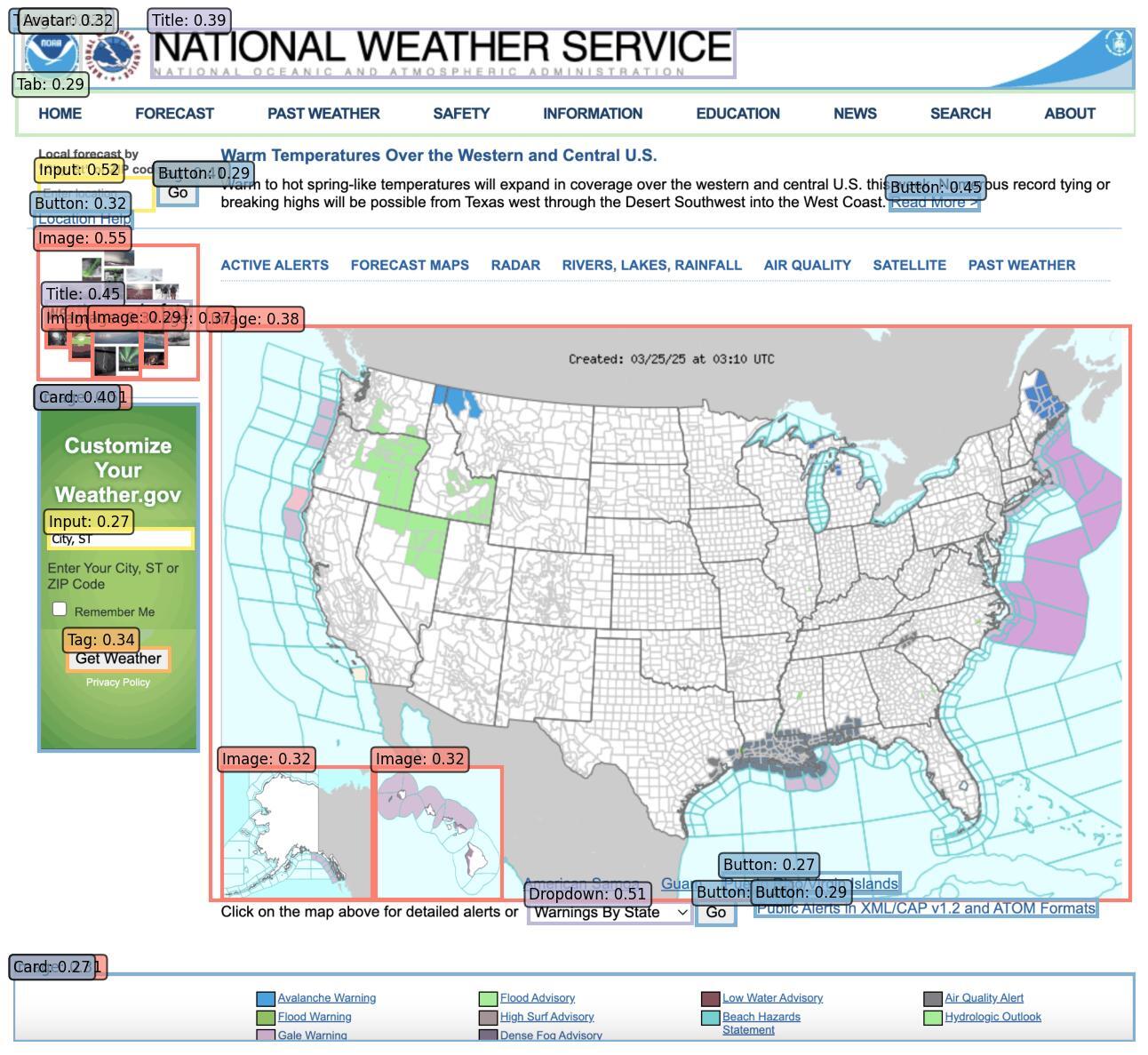}
    \caption{Example of visual element detection.}
    \label{fig:element_detection}
\end{figure}

\paragraph{Implementation.}
We fine-tune both models on a dataset of 12,000 UI screenshots collected from public design repositories for 36 epochs (batch size 16, $1024 \times 1024$ pixels, AdamW, learning rate=$2\times10^{-4}$). Data augmentation included Large-Scale Jitter (LSJ) and CopyPaste. This approach achieves a mAP@0.7512, effectively outperforming standalone DETR (0.7356) and YOLOv10 (0.7212). Detected elements are cropped and stored in a repository for downstream generation. Qualitative examples are visualized in Figure \ref{fig:element_detection}.

\subsection{Schema-Guided Code Generation} 

To bridge the gap between visual layout and executable code, DOne decouples structural reasoning from implementation details through a two-stage schema-guided synthesis.

\paragraph{Schema as Structural Intermediate Representation.} Directly mapping pixels to HTML often causes ``structural hallucinations,'' where VLMs misinterpret nesting relationships. To mitigate this, we generate a Layout Schema—a hierarchical JSON blueprint that serves as a spatial-to-logical bridge. Instead of treating the schema merely as context, we utilize it to explicitly ground the visual elements. By prompting the VLM with layout regions and element coordinates, we enforce a strict transformation where fine-grained assets (from detection) are logically nested within their parent semantic containers (from segmentation). This step fixes the DOM hierarchy before any code syntax is generated, effectively solving the layout ambiguity. 

\paragraph{Schema-Constrained Code Synthesis.} In the final stage, the generated Schema acts as a structural skeleton for code synthesis. The VLM is tasked with ``fleshing out'' this skeleton with Tailwind CSS styling and integrating the specific visual assets. This decomposition ensures that the VLM focuses on styling and asset integration (e.g., selecting transparent icons vs. content images) without the burden of inferring the high-level layout, thereby guaranteeing both structural correctness and visual fidelity. We summarize the complete end-to-end process in the algorithm \ref{alg:pipeline}. 

\begin{algorithm}[t]
\caption{End-to-End DOne Pipeline}
\label{alg:pipeline}
\small
\begin{algorithmic}[1]
\STATE \textbf{Input:} Image $I$; \textbf{Output:} UI code $H$
\STATE $segments \gets \text{SegmentationModel}(I)$
\STATE $segments \gets \text{BoxOptimization}(segments, 0.2)$
\STATE $elements \gets \text{UIDetectionModel}(I)$
\STATE $p_{sch} \gets \text{BuildSchemaPrompt}(segments, elements)$ 
\STATE $schema \gets \text{VLM}(p_{sch}, I, segments, elements)$    
\STATE $structure \gets \text{ParseJSON}(schema)$
\STATE $p_{html} \gets \text{BuildHTMLPrompt}(structure, elements)$
\STATE $H \gets \text{VLM}(p_{html}, I, elements)$
\end{algorithmic}
\end{algorithm}

\section{Experiments}

\begin{table*}[t]
    \centering
    \begin{minipage}{0.57\textwidth} 
        \centering
        \resizebox{\linewidth}{!}{ 
            \begin{tabular}{lcccccc}
            \toprule
            & \multicolumn{2}{c}{Claude-3.7-Sonnet} & \multicolumn{2}{c}{Gemini-2.5-Pro} & \multicolumn{2}{c}{Qwen-VL-Max} \\
            \cmidrule(lr){2-3} \cmidrule(lr){4-5} \cmidrule(lr){6-7}
            Method & CLIP & GPT & CLIP & GPT & CLIP & GPT \\
            \midrule
            S2C & 0.6475 & - & 0.6375 & - & 0.6131 & - \\
            CoT2C & 0.6499 & 0.625 & 0.6362 & 0.451 & 0.6076 & 0.487 \\
            DCGen & 0.6732 & 0.662 & 0.6553 & 0.487 & 0.6332 & 0.502 \\
            DOne & \textbf{0.7435} & \textbf{0.723} & \textbf{0.6837} & \textbf{0.554} & \textbf{0.6714} & \textbf{0.607} \\
            \bottomrule
            \end{tabular}
        }
        \caption{Performance comparison on HiFi2Code. GPT Score is pairwise preference vs. S2C.}
        \label{tab:backbone_comparison}
    \end{minipage}
    \hfill
    \begin{minipage}{0.4\textwidth} 
        \centering
        \resizebox{\linewidth}{!}{ 
            \begin{tabular}{lcccc} 
                \toprule
                Method & Text & Block & Pos & Color \\
                \midrule
                S2C & 0.6231 & 0.2942 & 0.561 & 0.051 \\
                CoT2C & 0.6155 & 0.2987 & 0.545 & 0.031 \\
                DCGen & 0.6454 & \textbf{0.3231} & 0.557 & 0.061 \\
                DOne & \textbf{0.6685} & 0.3088 & \textbf{0.572} & \textbf{0.461} \\
                \bottomrule
            \end{tabular}
        }
        \caption{Fine-Grained measurements using the same Qwen-VL-Max backbone.}
        \label{tab:subset_metrics}
    \end{minipage}
\end{table*}

\subsection{Experimental Setup}

To ensure a rigorous and fair evaluation, we conduct all experiments under a unified setting, controlling for backbone models, random seeds, and decoding parameters. For reproducibility, all source code, datasets, and other metadata are described in the Appendix~\ref{app:data_availability}.

\paragraph{Datasets.}
HiFi2Code: HiFi2Code comprises 200 high-fidelity webpage screenshots, rigorously curated for visual richness and structural complexity, featuring high element density and deep nesting hierarchies.

\paragraph{Backbones.} We employ Claude-3.7-Sonnet \cite{claude}, Gemini-2.5-Pro \cite{gemini}, and Qwen-VL-Max \cite{qwen} to test generalization across VLMs. For reproducibility, we utilize API versions and set the parameters (e.g., temperature) as default to ensure deterministic outputs.

\paragraph{Baselines.} We compare DOne against the three representative open-source methods:
(1) Screenshot2Code (S2C), representing standard end-to-end VLM generation \cite{Screenshot2Code} \cite{Screenshot2Code}.
(2) CoT2C, which leverages Chain-of-Thought reasoning before generation \cite{wei2022chain}.
(3) DCGen: a strong baseline utilizing a divide-and-conquer mechanism \cite{DCGen}.

\paragraph{Evaluation Metrics.}

To comprehensively assess performance from visual, textual, and structural perspectives, we employ a multi-level metric suite:

\textbf{High-Level Similarity}: We report the \textbf{CLIP Score} to measure the visual embedding similarity between the rendered output and the ground truth. We also utilize a \textbf{GPT-4o Judge} (GPT Score, see Appendix~\ref{app:gpt_score_judge_prompt}) for pairwise assessment.

\textbf{Fine-Grained Measurements}: 
To address the lack of detail in high-level metrics, we adopt the element-matching suite from prior work \cite{Design2Code}. We compute: \textbf{Block Match} (ratio of matched block sizes), \textbf{Text Match} (character overlap via Sørensen-Dice coefficient), \textbf{Color Match} (perceptual difference via CIEDE2000), and \textbf{Position Match} (center alignment accuracy).

\textbf{Human Evaluation}: We conduct a human evaluation with professional developers to judge visual fidelity and code usability.

\begin{table}[t]
    \centering
    \begin{tabular}{lcc}
        \toprule
        Configuration & CLIP & GPT \\
        \midrule
        Direct Generation & 0.6475 & - \\
        + Layout Segmentation & 0.7226 & 0.701 \\
        + Element Detection & 0.6975 & 0.656 \\
        + Schema Guidance & 0.6615 & 0.554 \\
        + All Techniques (\textbf{DOne}) & \textbf{0.7435} & \textbf{0.723} \\
        \bottomrule
        \end{tabular}
    \caption{Ablation study on DOne components.}
    \label{tab:ablation}
\end{table}

\subsection{Main Results}

\subsubsection{Performance on HiFi2Code}
\paragraph{High-Level Similarity Analysis.} Table \ref{tab:backbone_comparison} presents the comparison results. DOne consistently outperforms all baselines across all backbones.
Notably, on the powerful Claude-3.7-Sonnet, DOne achieves a CLIP score of 0.7435 and a GPT Score of 0.723, surpassing the previous DCGen by significant margins (approx. +10\% in GPT Score).
This improvement indicates that our learned layout segmentation and element retrieval provide stronger grounding than heuristic cropping (DCGen) or pure reasoning (CoT2C). The low performance of S2C highlights the necessity of the hierarchical approach for complex pages.

\paragraph{Fine-Grained Structural Analysis.} 

To complement high-level semantic metrics, we conduct a rigorous structural analysis on a subset of 50 randomly selected webpages from HiFi2Code. 
Direct comparison against raw production code is flawed due to obfuscation and framework-specific bloat (e.g., React wrappers). To enable a semantic comparison, we utilize a ``Canonical Ground Truth'' (Appendix~\ref{app:appendix_gt_construction}). This provides a clean, standardized HTML code for metric computation.

As shown in Table~\ref{tab:subset_metrics}, while baselines like DCGen achieve competitive Block Match scores (0.3231) by correctly placing generic placeholders, they fail catastrophically in Color Match ($<0.07$), indicating a lack of fine-grained styling.
In contrast, DOne achieves a Color Match of 0.4612 and a superior Text Match, confirming that our schema-guided approach retrieves and renders actual visual assets rather than mere wireframes.
DOne also leads in Position Match (0.5721) and Text Match (0.6685), indicating that our schema-guided generation anchors elements more accurately in the DOM tree than Chain-of-Thought and heuristic methods.

\subsection{Ablation Study}
To assess the contribution of each module (layout segmentation, visual element detection, and hierarchical schema generation.), we conduct an ablation study using Claude-3.7-Sonnet under identical seeds and decoding parameters.

As shown in Table \ref{tab:ablation}, introducing Layout Segmentation provides the most significant gain (+0.0751 CLIP, 0.701 GPT Score), confirming that macro-structure decomposition is critical for reducing layout errors.
Element Detection alone improves fine-grained fidelity (CLIP 0.6975) but lacks structural coherence.
Schema Generation alone offers limited gains (CLIP 0.6615), indicating that structural priors require visual grounding.
The full DOne framework achieves the highest performance (CLIP 0.7435, GPT Score 0.723), demonstrating that the synergy of layout decomposition, element retrieval, and schema guidance is essential for high-fidelity reconstruction.

\subsection{Human Evaluation}
While automated metrics provide a proxy for quality, practical utility requires human assessment. We followed prior protocols~\cite{Design2Code} to conduct a user study, rigorously addressing experimental control (blinding, shuffling) to ensure validity.

\subsubsection{Visual Fidelity}

\paragraph{Direct Comparison.} To quantitatively assess visual fidelity from an end-user perspective, we conduct an online questionnaire with 20 developers and designers. The evaluation covers 50 randomly sampled real-world webpages.
For each test case, the generated outputs from S2C, CoT2C, DCGen, and DOne are presented in an anonymized and randomized order to mitigate evaluator bias. Participants are asked to select the best result that replicated the reference webpage.
As illustrated in Table~\ref{fig:visual_sim}, DOne secured the highest preference, with 58.6\% of evaluators identifying it as the closest match to the original design. In contrast, CoT2C, S2C, and DCGen received only 23.3\%, 15.5\%, and 2.6\% of the votes, respectively. These results indicate that DOne achieves superior visual fidelity.

\begin{table}[t]
    \centering
    \small
    \begin{tabular}{lc}
        \toprule
        Method & Vote Percentage \\
        \midrule
        S2C & 15.5\% \\
        CoT2C & 23.3\% \\
        DCGen & 2.60\% \\
        DOne & \textbf{58.6\%} \\
        \bottomrule
    \end{tabular}
    \caption{User preference distribution.}
    \label{fig:visual_sim}
\end{table}

\begin{figure}[t]
    \centering
    \includegraphics[width=\linewidth]{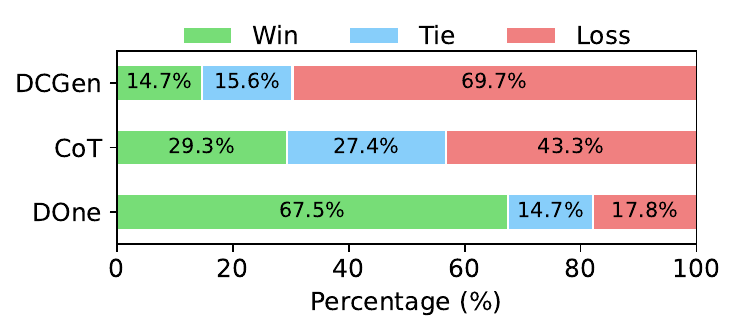}
    \caption{Pairwise win/tie/loss rates vs. S2C.}
    \label{fig:pair_wise}
\end{figure}

\begin{table}[t]
    \centering
    \small
    \begin{tabular}{lcc}
        \toprule
        Metric & Control & DOne (Exp) \\
        \midrule
        Time (min) & 9.81 & \textbf{3.18} \\
        CLIP Score & 0.499 & \textbf{0.665} \\
        \bottomrule
    \end{tabular}
    \caption{Average efficiency and fidelity metrics.}
    \label{tab:efficiency}
\end{table}

\begin{figure}[t]
    \centering
    \includegraphics[width=0.9\linewidth]{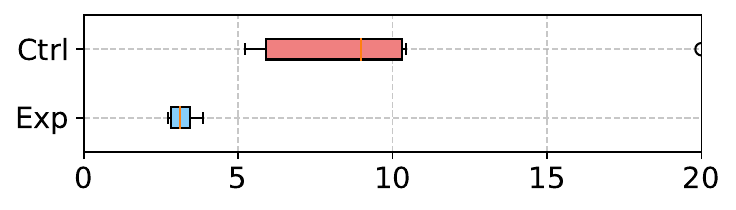}
    \caption{Task completion time distribution.}
    \label{fig:time}
\end{figure}

\paragraph{Pairwise Comparison.} Complementing the direct preference survey, we also conduct a pairwise evaluation to assess the relative improvements of DOne. Five participants compare the generated outputs across the same 50 webpages. For each test case, participants are presented with the reference image and two generated outputs: one from the S2C baseline and one from a target method (DOne, DCGen, or CoT2C). Participants determine which output better preserved visual structure and details, with final outcomes (Win/Tie/Loss) established via majority consensus ($\ge 3$ votes). As shown in Figure \ref{fig:pair_wise}, DOne outperforms the baseline in 67.5\% of cases while losing in only 17.8\%, demonstrating its superiority. A one-sided Binomial test yields a $p$-value of 0.0077 ($p < 0.01$), achieving statistical significance.

\paragraph{Discussion.} It is noteworthy that certain discrepancies exist between automated metrics and human judgment. This inconsistency highlights the limitations of current automated metrics in evaluating UI code generation tasks. Specifically, while DCGen achieves a reasonable CLIP score (Table \ref{tab:backbone_comparison}), it received markedly low human ratings. Qualitative feedback suggests DCGen captures semantic content but fails in precise element rendering (e.g., distorted icons, misalignment). Human evaluators penalize these local visual artifacts heavily, whereas global metrics like CLIP are less sensitive to fine-grained structural fidelity.

\subsubsection{Efficiency Evaluation}
To quantify real-world productivity gains, we conduct a controlled within-subjects experiment with six participants. 

\paragraph{Experimental Protocol.} 
Participants are tasked with implementing web pages from screenshots within a 20-minute limit. We employ a counter-balanced design where each participant used both the DOne-assisted workflow and their preferred AI-assisted toolchain (e.g., Cursor with Claude-3.7-Sonnet) as a control. A task is recorded as ``complete'' only when the generated code is syntax-error-free and visually aligned with the reference.

\paragraph{Results.} 
As shown in Table~\ref{tab:efficiency} and Figure~\ref{fig:time}, DOne demonstrates significant efficiency improvements, reducing the average completion time from 9.81 min to 3.18 min (a $3\times$ speedup). 
In terms of quality, DOne achieves superior visual fidelity. The average CLIP score for the DOne group is 0.665, significantly outperforming the control group's 0.499. 
While the control group required extensive manual debugging just to meet the completion standards, they still resulted in lower visual fidelity. 
In summary, participants using DOne achieves faster and more accurate UI code translation, demonstrating its utility in accelerating development.

\section{Conclusion}
In this work, we propose DOne, a novel framework designed to overcome the ``holistic bottleneck'' inherent in monolithic VLMs. By decoupling layout segmentation from fine-grained element retrieval and grounding generation in an intermediate schema, DOne effectively harmonizes global structural coherence with local visual details. 
To foster research in this field, we release WebSeg to facilitate robust layout learning and HiFi2Code to establish a high-fidelity evaluation benchmark aligning with real-world production standards.
Extensive evaluations demonstrate that DOne significantly outperforms existing methods in both visual fidelity and developer productivity.

\section*{Limitations}

Our current element retriever relies on a fixed taxonomy of UI components. Future work could explore open-vocabulary detection to handle novel design trends. Additionally, while we support static visual assets, complex dynamic interactions (e.g., JS animations) remain a challenge for future research.


\bibliography{arxiv/DOne}

\appendix

\section{GPT Score Judge Prompt}
\label{app:gpt_score_judge_prompt}

The prompt provided in Figure \ref{fig:gpt_score_judge_prompt} is utilized to conduct the pairwise preference evaluation (GPT Score). We instruct the model to compare two generated candidates (Method A and Method B) against the original reference image, specifically prioritizing layout accuracy and fine-grained visual details in its decision-making process.
\begin{figure}[t]
    \centering
    \fbox{
    \begin{minipage}{0.9\linewidth}
        \vspace{0.5em}
        You are an expert web UI evaluator. I'm showing you three images:
        
        \begin{itemize}
            \item Image 1: The original web UI design reference
            \item Image 2: Method A generated UI
            \item Image 3: Method B generated UI
        \end{itemize}
        
        Please compare both methods (A and B) with the original design and determine which one is better at reproducing the original design.
        
          Consider these aspects in your evaluation:
          \begin{enumerate}
              \item Visual similarity to the original
              \item Layout accuracy
              \item Component positioning
          \end{enumerate}
        
          Make your final decision based on which method looks more similar to the original design.

        Format your response as follows:
        \begin{quote}
            \texttt{evaluation\\
            WINNER: [METHOD A or METHOD B]\\
            REASONING: [...]}
        \end{quote}
        \vspace{0.5em}
    \end{minipage}
    }
    \caption{GPT Score Judge Prompt.}
    \label{fig:gpt_score_judge_prompt}
\end{figure}

\section{Construction of Canonical Ground Truth}
\label{app:appendix_gt_construction}

For the fine-grained structural analysis on the subset ($N=50$), relying on the original raw source code of webpages is problematic. Real-world production code is often heavily obfuscated, relies on complex frameworks (e.g., React, Vue), or contains redundant legacy CSS, making it unsuitable for direct DOM-based metric comparison (e.g., Block Match or Tree matching) against model-generated clean HTML.

To address this, we construct a set of ``Canonical Ground Truth'' files (\texttt{gt.html}) that are visually identical to the reference screenshots but implemented in clean, standard HTML/Tailwind CSS. We employ a two-stage high-fidelity reconstruction pipeline:

\begin{enumerate}
    \item Visual Parsing via Qwen-2.5-VL: We utilize Qwen-2.5-VL for its state-of-the-art visual grounding capabilities. The model was prompted to analyze the screenshot and output a structured description, including precise bounding box coordinates for all layout containers, OCR text content, and element types (buttons, inputs, images).
    
    \item Code Synthesis via Gemini-3.0: The structured visual parsing results were fed into Gemini-3.0. The model was tasked with synthesizing the final \texttt{gt.html} code using the provided coordinates and text to ensure pixel-level structural alignment, while normalizing the styling into standard Tailwind CSS.
\end{enumerate}

Finally, we perform a manual verification on the generated \texttt{gt.html} files to ensure they strictly adhered to the visual appearance of the original HiFi2Code screenshots. These standardized ground truth files serve as the reference for the Text, Block, Position, and Color match metrics.

To ensure this GT remains a neutral baseline, we implement three safeguards: (1) Implementation-Agnostic Metrics: Evaluations are based on rendered visual properties (e.g., computed RGB values, bounding boxes) rather than source code syntax, treating Tailwind and raw CSS identically; (2) Visual-Centric Verification: The GT underwent strict human verification solely for pixel-level visual alignment, independent of coding schema; (3) Empirical Validation: DOne does not dominate structural metrics like Block Match compared to baselines (which also generate Tailwind CSS), explicitly contradicting the hypothesis of structural bias.

\section{Prompt Template}
\label{app:prompt}

This appendix provides the prompt templates used to guide the VLMs in the two-stage code generation process. The first prompt instructs the model to produce a structured layout schema, while the second directs the synthesis of pixel-accurate HTML and Tailwind CSS code.

\begin{figure*}[htbp]
    \centering
    \includegraphics[width=0.85\textwidth]{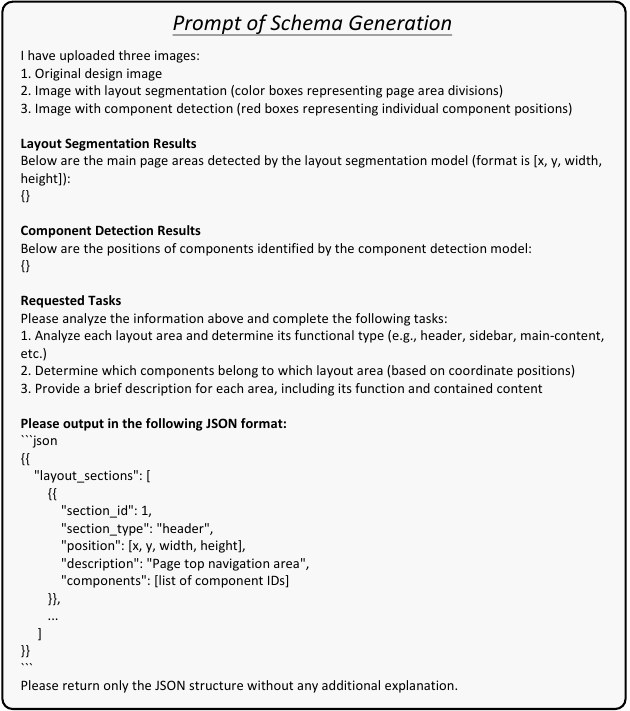} 
    \caption{Prompt of schema generation.}
    \label{fig:prompt_schema}
\end{figure*}

The prompt displayed in Figure~\ref{fig:prompt_schema} is designed to elicit a hierarchically structured JSON representation of the input design. It directs the VLMs to analyze the provided layout segmentation and component detection results, assign functional types to each layout region (e.g., header, sidebar), associate detected elements with their parent sections based on spatial coordinates, and describe the purpose and content of each area. The output is a structured schema that serves as an intermediate representation for the subsequent code generation stage.

\begin{figure*}[htbp]
    \centering
    \includegraphics[width=0.95\textwidth]{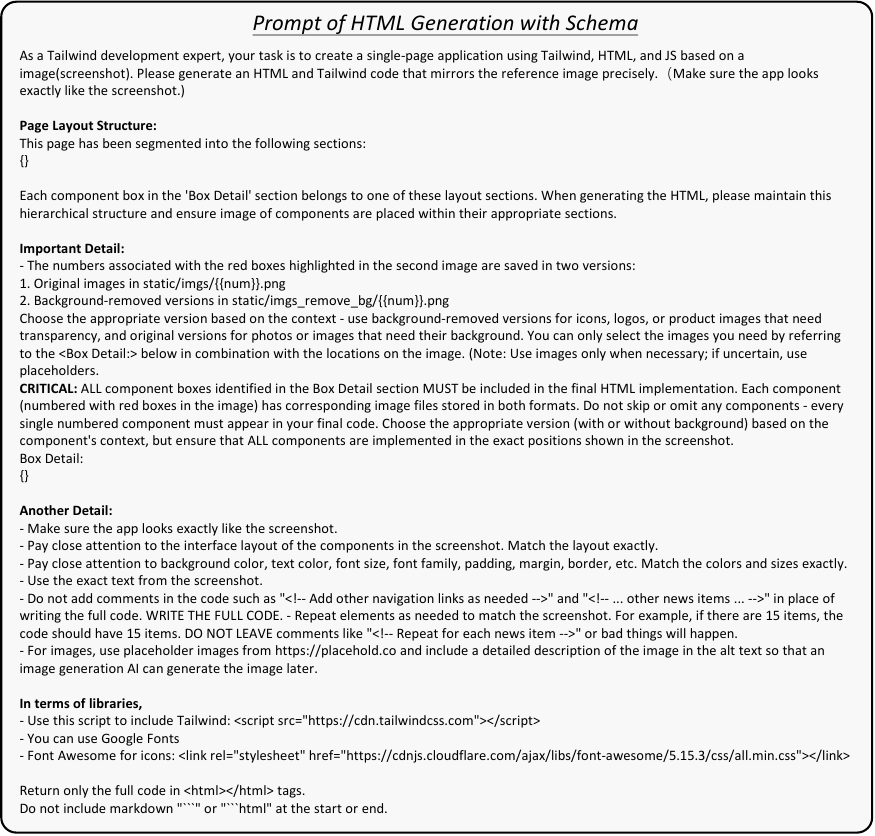} 
    \caption{Prompt of code generation with schema.}
    \label{fig:prompt_html_schema}
\end{figure*}

The prompt shown in Figure~\ref{fig:prompt_html_schema} guides the VLMs to generate production-ready, Tailwind CSS-based HTML code. It incorporates the layout schema from the previous stage and provides detailed instructions for achieving high visual fidelity. Key directives include: strict adherence to the original screenshot's layout and styling, appropriate use of pre-extracted visual assets (in both original and background-removed versions), prohibition of placeholder comments, and precise replication of all visual elements, text content, and styles. The prompt emphasizes functional completeness and pixel-perfect accuracy in the final output.

\section{Example Results of Layout Segmentation and Element Detection}

To illustrate the types of inputs provided to the model, consider the following examples of layout segmentation and component detection outputs that would be incorporated into the prompt.

\paragraph{Example of Layout Segmentation Results}
The layout segmentation model identifies major structural regions within the interface. Each region is represented as a bounding box with coordinates and a confidence score, as shown in the following example output:

Below are the main page areas detected by the layout segmentation model (format is [x, y, width, height]):
\begin{verbatim}
[
  {
    "category_id": 0,
    "bbox": [0, 0, 2770, 220],
    "score": 1.0
  },
  {
    "category_id": 0,
    "bbox": [0, 223, 2270, 126],
    "score": 1.0
  },
  {
    "category_id": 0,
    "bbox": [0, 385, 1930, 925],
    "score": 1.0
  },
  {
    "category_id": 0,
    "bbox": [1930, 401, 848, 925],
    "score": 1.0
  }
]
\end{verbatim}

\paragraph{Example of Visual Element Detection Results}
The component detection model identifies individual UI elements within the design, providing precise coordinates for each detected component. 
Below are the positions of components identified by the component detection model:
\begin{itemize}
  \item Box2: $x=846,\; y=50,\; \text{width}=210,\; \text{height}=100$
  \item Box4: $x=700,\; y=400,\; \text{width}=480,\; \text{height}=365$
  \item Box6: $x=1962,\; y=403,\; \text{width}=105,\; \text{height}=105$
  \item Box8: $x=2268,\; y=0,\; \text{width}=267,\; \text{height}=48$
\end{itemize}

\section{Model Performance}
This section presents the detailed quantitative results of the layout segmentation and visual element detection models described in Sections \ref{sec:layout_section} and \ref{sect:element_section}, providing empirical validation of their effectiveness in the proposed workflow.

\subsection{Quantitative Results of Layout Segmentation Model}
The performance of the RT-DETR-based layout segmentation model, whose architecture and training configuration are detailed in Section \ref{sec:layout_section}, is quantitatively evaluated across standard detection metrics. The results presented in Table \ref{tab:seg_model} demonstrate the model's strong capability in accurately identifying and localizing macroscopic layout regions within web interface designs.

\subsection{Comparative Performance of Element Detection Model}
To validate the hybrid detection approach introduced in Section \ref{sect:element_section}, we compare the performance of the individual models and their combination across different object size categories. Table \ref{tab:map_comparison} presents the mAP$_{75}$ scores for large objects, small objects, and overall performance, confirming that the integrated CO-DETR+YOLOv10 framework achieves superior performance by leveraging the complementary strengths of both architectures.

\begin{table}[t]
\centering
\caption{Performance metrics}
\begin{tabular}{lc}
\toprule
Metric & Value \\
\midrule
mAP$_{50:95}$ & 0.5316 \\
mAP$_{50}$    & 0.8203 \\
Precision     & 0.7763 \\
Recall        & 0.7293 \\
\bottomrule
\end{tabular}
\label{tab:seg_model}
\end{table}

\begin{table*}[htbp]
\centering
\caption{Comparison of mAP$_{75}$ across different model combinations.}
\begin{tabular}{lccc}
\toprule
Model & Large Objects & Small Objects & Overall \\
\midrule
CO-DETR & \textbf{0.6823} & 0.7545 & 0.7356 \\
YOLOv10 & 0.6139 & \textbf{0.7914} & 0.7212 \\
CO-DETR+YOLOv10 & 0.6751 & 0.7812 & \textbf{0.7512} \\
\bottomrule
\end{tabular}
\label{tab:map_comparison}
\end{table*}

\section{Case Visualization}
\label{app:case_visual}

\begin{figure*}[t]
    \centering
    \includegraphics[width=0.9\textwidth]{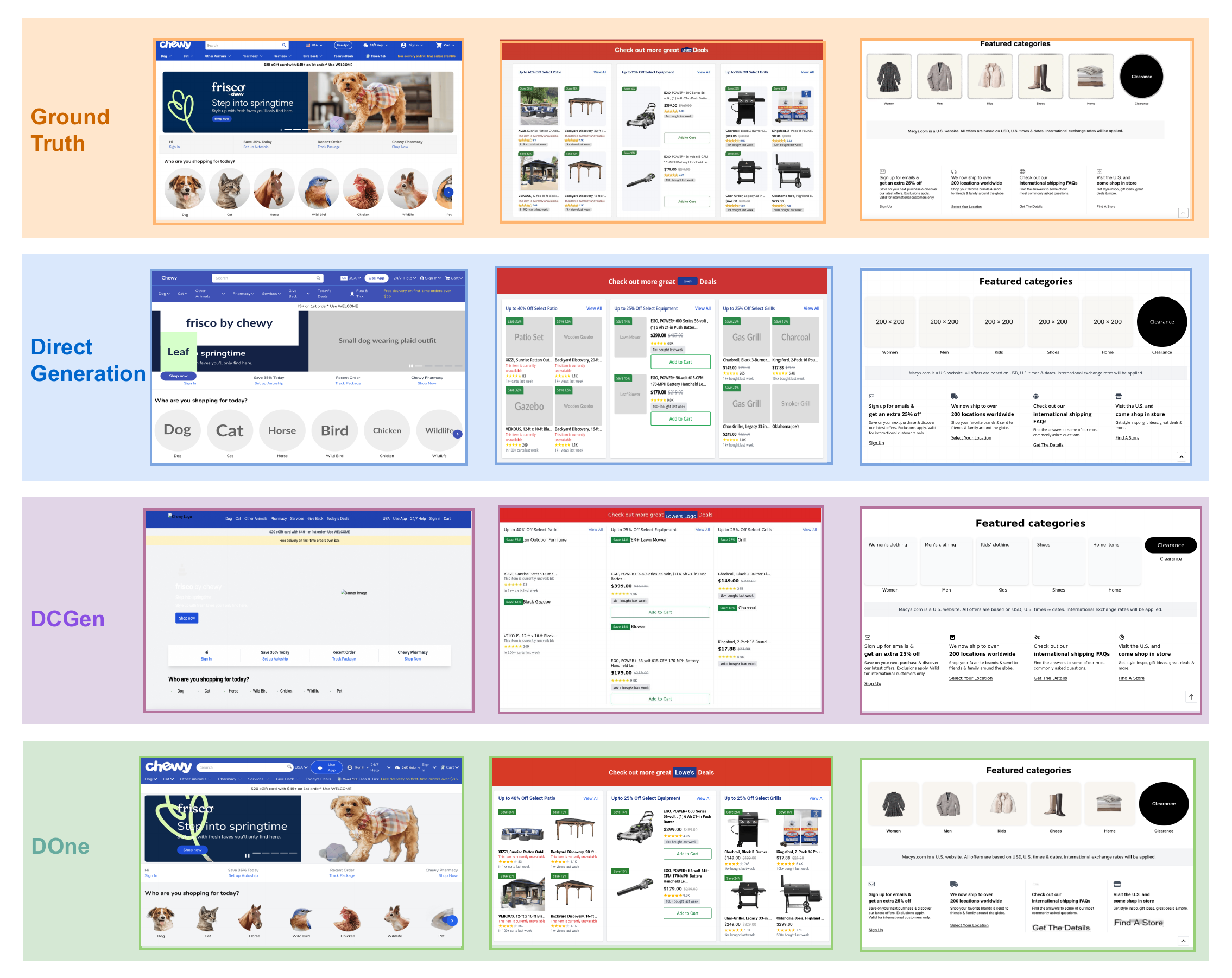} 
    \caption{Qualitative comparison of generated webpage results.}
    \label{fig:case_study}
\end{figure*}

To provide a more intuitive and qualitative assessment of DOne's capabilities, we present a series of case studies in Figure \ref{fig:case_study}, comparing the outputs of our DOne against the ground truth and a standard direct generation baseline. The figure clearly illustrates that DOne consistently produces front-end code that is significantly more faithful to the original webpage design. For instance, existing methods fail to render critical visual components, such as the main promotional image and the specific product images, often substituting them with incorrect content (e.g., ``Leaf Springtime'') or generic placeholders (e.g., ``Men'', ``Kids''). In stark contrast, DOne leverages its component detection module to accurately identify and extract these key elements—from hero images to individual items—and correctly embeds their information into the generated code, resulting in a visually and functionally accurate rendition.

Furthermore, the case studies highlight DOne's superior ability to handle complex page structures, particularly those with multiple elements and dense content. In Case 2, which features a complex grid layout of products, the DCGen method struggles with alignment and structural integrity, leading to a disorganized and visually jarring result. Our DOne, however, benefits from its layout segmentation capability, allowing it to correctly parse the complex arrangement of the page. It successfully preserves the intended grid structure, element spacing, and overall coherence. This demonstrates that DOne not only recognizes individual components but also understands their spatial relationships and the overarching layout. The synergy between component detection and layout segmentation enables DOne to robustly reconstruct webpages with high fidelity, tackling challenges where simpler methods fall short.

\end{document}